\documentclass[11pt,a4paper]{article}
\usepackage[hyperref]{naaclhlt2018}
\usepackage{times}
\usepackage{latexsym}

\usepackage{url}

\usepackage{latexsym}
\usepackage{url}
\usepackage[toc,page]{appendix}
\usepackage{titlesec}
\usepackage{amsmath,amssymb}
\usepackage{algpseudocode}
\usepackage[]{algorithm2e}
\usepackage{geometry}
\usepackage{amsfonts}
\usepackage[flushleft]{threeparttable}
\usepackage{tablefootnote}
\usepackage{subcaption}
\usepackage{tabularx}
\usepackage{hyperref}
\usepackage{fancyhdr}
\usepackage{longtable}
\usepackage{url}
\usepackage{leftidx}
\usepackage{graphicx,grffile}
\usepackage{multirow,bigstrut}
\usepackage{soul,color}
\usepackage{perpage} 
\usepackage{booktabs}

\MakePerPage{footnote}

\aclfinalcopy 

\newcommand\norm[1]{\left\lVert#1\right\rVert}

\newcommand{\argmax}[1]{\underset{#1}{\text{arg}\,\text{max}}\;}

\title{Graph Convolutional Networks for Classification \\ with a Structured Label Space}

\author{Meihao Chen \\
  Bombora Inc. \\
  {\tt mchen@bombora.com} \\\And
  Zhuoru Lin \\
  Center for Data Science, \\New York University  \\
  {\tt zlin@nyu.edu} \\\And
  Kyunghyun Cho \\
  Center for Data Science, \\New York University  \\
  {\tt kyunghyun.cho@nyu.edu} \\}

\date{}

\begin{document}
\maketitle
\begin{abstract}

  It is a usual practice to ignore any structural information underlying classes in multi-class classification. In this paper, we propose a graph convolutional network (GCN) augmented neural network classifier to exploit a known, underlying graph structure of labels. The proposed approach resembles an (approximate) inference procedure in, for instance, a conditional random field (CRF). We evaluate the proposed approach on document classification and object recognition and report both accuracies and graph-theoretic metrics that correspond to the consistency of the model's prediction. The experiment results reveal that the proposed model outperforms a baseline method which ignores the graph structures of a label space in terms of graph-theoretic metrics.
\end{abstract}

\section{Introduction}

Multi-class classification is one of the most common
problems in machine learning. It aims at predicting one label out of multiple, mutually exclusive labels based on the known assignments in a training data.

Such an approach does not take into account complex dependencies among output variables, potentially leading to two problems.
First, it assumes mutually independent labels. The assumption holds on some computer vision tasks such as object recognition on ILSVRC \citep{ILSVRC15}, in which classes are mutually exclusive leaf nodes of WordNet \citep{wordnet} 
(e.g., an object is not supposed to be both a dog and of a cat), but does not apply to many other tasks. 
Second, the quality of top-k predictions is not well-assessed. 
A naive multi-class classification framework evaluates top-k accuracy, which only measures the model's ability to exactly match the true label and ignores the relevance of other top predictions. 
This is critical especially in a dataset with highly correlated classes. 
For example, an image labeled with `husky' is classified as `dog' or `mammal'.
Though neither matches the ground truth exactly, `dog' is clearly a better prediction than `mammal'.
Top-k predictions of `dog' \& `husky' should be considered better than that of `mammal' \& `husky'. 

A known label relation can be exploited as a guide for a model to produce a cluster of predictions that are close to the ground truth in a structured label space. As a result, both classification accuracy and the relevance of top predictions can be improved. Graphs have been shown to encode a complex geometry and can be used with strong mathematical tools such as spectral graph theory \citep{chung1997spectral}. 

There have been work on incorporating the label structure in multi-class classification. They however come with two major shortcomings.
First, classification with label relations is often confined to a certain type of graph \citep{deng2014large}. However, underlying label relations of a certain task may exist in various ways.
Second, most of the recent work approximates the pairwise relation with graphical models such as conditional random fields (CRF) and Markov random fields (MRFs) \citep{DBLP:journals/corr/SchwingU15}. These may not be rich enough to capture complex dependencies.

In this paper, we explore a novel way to perform multi-class classification combining deep neural networks (DNN) with graph convolutional networks (GCN) \citep{DBLP:journals/corr/BrunaZSL13} that encodes the label structure and improves top predictions relevancies.
The proposed model stacks graph convolution layers on the concatenation of input and class latent variables to extract label-wise features, which are then decoded by a final classifier. The entire network is trained as a deterministic deep neural network, bypassing any need for sophisticated inference steps. 
We also propose several graph-theoretic metrics to evaluate the relevancy of top predictions.

\section{Problem Description}


Given an input variable $x \in \mathcal{X}$, and output variables $\mathcal{Y} = \left\{1, \ldots, L \right\}$,
the classification task amounts to assigning $x$ with an output variable $y_i$ that maximize the probability. 
When correlations exist among output variables, the probabilities of a certain $y_i$ depends on not only $x$, but also the output variables that $y_i$ has correlations with. One way to represent such known structure underlying the classes is to use graph structure. Let $\mathcal{G} = (V, E)$ be a graph such that $y$ is indexed by the vertices of $\mathcal{G}$, and $e_{ij} \in E$ is an edge between $y_i$ and $y_j$ that represents a known relation between the output random variables, the problem amounts to modelling the probability $p(y|x, \mathcal{G})$.

\section{Class Structure Aware Classification}

\subsection{General Setup}

The goal of supervised learning is to map an input $x$ to one of the classes $\mathcal{Y}$. This process often consists of three sub-modules. The first module extracts {\bf input representation}, and the second module {\bf class representation}s. The final module, called a {\bf score function}, compares the input representation against each of the class representation to compute the score $F(x, y; \theta)$ of each class given the input. Given the scores, the prediction is made by
$\hat{y} = \argmax{y \in \mathcal{Y}} F(x, y; \theta),$ where $\theta$ denotes a set of parameters of the classifier.

With the score function above, we define a conditional distribution over the classes given an input. This is often done by so-called softmax:
\[
p(y|x; \theta) = \frac{\exp(F(x, y; \theta))}{\sum_{y' \in \mathcal{Y}} \exp(F(x,y';\theta))}.
\]
With this conditional distribution, we can now maximize a log-likelihood given a set $D$ of training examples with respect to the parameters:
$\hat{\theta} = \argmax{\theta} \sum_{(x, y^*) \in D} \log p(x, y^*; \theta).$

\paragraph{Example: Multilayer Perceptron (MLP)}

When it is assumed that each class is conditionally independent from each other and that there is no known structure underlying the classes, we can use a plain multilayer perceptron (MLP). First, the input representation $f(x; \theta_f) \in \mathbb{R}^d$ is extracted by a deep neural network. The representation of each class $y$ is simply a trainable vector $v_y \in \mathbb{R}^d$ and does not depend on the other classes nor on the input. The score function is a dot product between the input representation and the class vector, i.e., $F(x, y; \theta) = v_y^\top f(x)$, where $\theta = \theta_f \cup \left\{ v_1, \ldots, v_{L} \right\}$.

\subsection{Structured Class Space}

In this paper, we are interested in the case where there exists a graph structure underlying the classes in $\mathcal{Y}$. This graph $\mathcal{G}$ indicates the similarity or relatedness between each pair of classes. The degree of similarity between the classes $i$ and $j$ is given by the weight $a_{ij}$, and these weights collectively define an adjacency matrix $A = \left[ a_{ij} \right]_{|\mathcal{Y}| \times |\mathcal{Y}|}$. In this paper we focus on undirected graph, therefore $A$ is symmetric.

\paragraph{Example: MLP + CRF}
Instead of defining the score function as a dot product between the input and class representations, we consider a conditional random field defined over the classes $\mathcal{Y}$ given an observation $x$ with a unary potential function $\psi_i(x; \theta_i)$ modelled by MLP and a pairwise potential function $\psi_{ij}(y_i, y_j, x; \theta_{ij})$. The score associated with the $i$-th class consists of both unary and pairwise potentials:
\begin{align*}
F(y_i, x; \theta) = \psi_i(x; \theta_i) + \sum \limits_{j=1, j\neq i}^{N} \psi_{(i,j)}(y_i, y_j, x; \theta_{ij}).
\end{align*}

For a general graph, the problem of exact inference in CRFs is intractable. Instead, mean field (MF) inference can be used to obtain an approximate solution. Initializing the score function to be $F_0(y_i, x; \theta) = \psi_i(x; \theta_i)$, the score function of $y_i$ at iteration $t$ is:
\begin{equation}\label{eq:mf}
F_i^t(y_i, x; \theta) =\\
F_i^{t-1}(y_i, x; \theta_i) + \sum_{j\in \mathcal{N}_i}\theta_{ij}\langle F_j^{t-1}(y_j, x; \theta_j)\rangle, 
\end{equation}
where $\langle F_j(y_j, x)\rangle$ resembles a ``message'' sent from node $j$ to $i$.
Assuming each node is a binary, the marginal probability of $y_i$ can be obtained at convergence using softmax.

\paragraph{Conditional Ising Model} Inspired by \citet{DBLP:journals/corr/DingDMN15}, we add an Ising model on top of MLP and use this as one of the baselines. An Ising model has score function $F(\textbf{y})$ that takes into account local potentials $h_i y_i$ as well as pairwise potential $J_{ij}y_i y_j$:
$F(\textbf{y}) = \sum_{(i,j)\in \mathcal{G}} J_{ij}y_i y_j + \sum_{i=1}^{n}h_i y_i.$

When an MLP maps an input feature vector $\textbf{x}=[x_1, \ldots , x_n]$ to 
a label bias vector $\textbf{z}=[z_1, \ldots , z_n]$, the conditional probability is defined as:
\begin{equation}\label{eq:cond1}
p(\textbf{y}|\textbf{z}) \propto \prod_{i=1}^{n} \psi_i(y_i, z_i)\prod_{(i,j)\in \mathcal{G}}\psi_{ij}(y_i, y_j; \theta_{ij}),
\end{equation}
where $\psi_i(y_i, z_i) = \exp(-y_i z_i)$, and $\psi_{ij}(y_i, y_j)$ is the edge-specific potential function.

The pairwise energy function $E(y_i, y_j)$ is define as:
$
E(y_i, y_j; \theta) = -\theta_{ij} y_i y_j,\text{ where }y_i \in \left\{ -1, 1 \right\}.
$
$\theta_{ij}$ is the interaction parameter defined to be set to $\theta$ if $(i,j) \in \mathcal{E}$ and otherwise, $0$.
Setting $\psi_{ij}(y_i, y_j) \propto \exp(-E(y_i, y_j; \theta))$, we can further rewrite Eq. \ref{eq:cond1} as: 
\begin{equation}
p(\textbf{y}|\textbf{z})  \propto 
  \exp(\sum_{(i,j)\in\mathcal{G}}\theta_{ij}y_i y_j - \sum_{i=1}^n z_iy_i).
\end{equation}

\subsection{Graph-based Output Structure}
\subsubsection{Graph Convolutional Network}

A graph convolutional network (GCN) is defined according to a graph $\mathcal{G} = (V, E)$, where $V$ is a set of nodes, and $E = V \times V$ is a set of edges. The edge weights $a_{ij}$ for $e_{ij} \in E$ form an adjacency matrix $A$. The node representation $h_v \in \mathbb{R}^D$ for node $v$ is usually obtained by a neural network, and node representations for all nodes collectively form a feature matrix $H = [h_1; h_2; ...; h_{|V|}]$. The network takes as input $A$ and $H$, and generates a node-wise output feature matrix $U \in \mathbb{R}^{|V|\times D_U}$. Each layer of propagation can be written as a non-linear form:
$H^{(l+1)} = g(H^{(l)}, A),$
where $H^{(0)} = H$ and $U = H^{(L)}$. $L$ is the number of layers. 

We use the iteration rule by \citet{kipf2016semi}:
\begin{equation}\label{eq:gcn}
H^{(l+1)} = \sigma(\hat{A}H^{(l)}W^{(l)}), 
\end{equation}
where $W^{(l)} \in \mathbb{R}^{D_l\times D_{l+1}}$ is a trainable parameter, and $\sigma(\cdot)$ is a non-linear function such as $\tanh$. Here, $\hat{A} = \tilde{D}^{-\frac{1}{2}}\tilde{A} \tilde{D}^{-\frac{1}{2}}$, in which $\tilde{A} = A + I$ is the adjacency matrix with a self-connection, and $\tilde{D}$ is the diagonal node degree matrix of $\tilde{A}$. $A$ is normalized such that all rows sum to one
. As the entire network is designed to be differentiable end-to-end, all the parameters are estimated with gradient-based optimization.
\subsubsection{Proposed Approach}
\begin{figure*}
\begin{minipage}[c]{0.63\textwidth}
\centering
 \includegraphics[width=\linewidth]{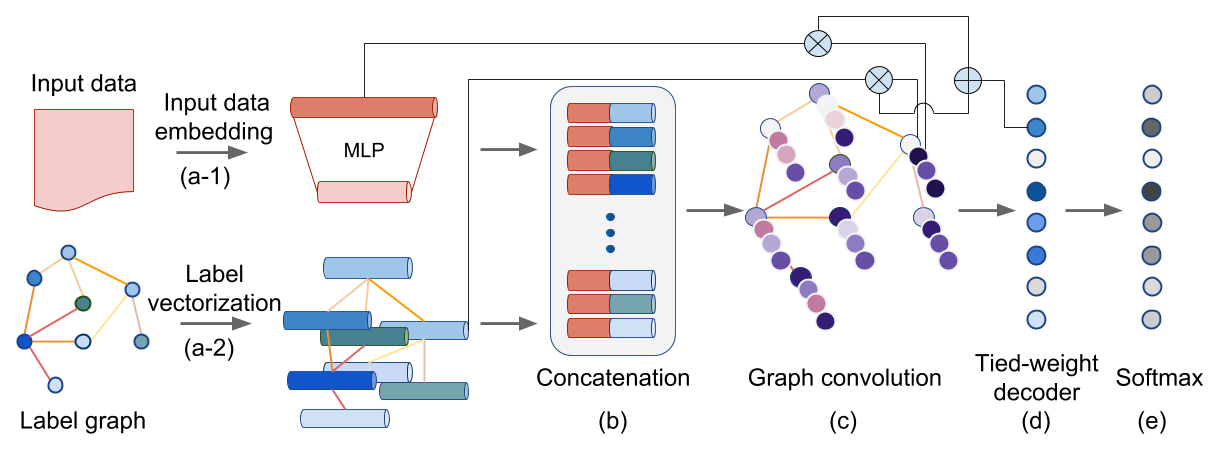}
 \end{minipage}
 \begin{minipage}[c]{0.36\textwidth}
    \caption{\small Model architecture: multi-class classification with label representation learning through GCN propagations. (a-1) Input data representation and MLP to generate context latent vector; (a-2) initialize label vectors; (b) concatenate each label vector with context vector to create conditional label representation; (c) graph feature extraction through GCN propagations; (d) tied-weight decoder; (e) softmax label scores to output probabilities.} \label{fig:structure}
\end{minipage}

\vspace{-8mm}
\end{figure*}

The neural message passing procedure with GCNs in Eq. \ref{eq:gcn} is similar to mean field iterations in Eq. \ref{eq:mf}, as each node is associated with a certain quantity that is computed based on the associated quantities of the neighboring nodes. 
The major difference between the two methods is that in GCN the value of each class node or label is a vector representation instead of a scalar. To compute the score of each label, the vector needs to be transformed into a scalar. We hereby introduce our methods of label representation and decoder. The input representation and GCN propagation remain the same as previous sections.

\paragraph{Context-Dependent Label Representation} 

Following the MLP example, the input representation $z = f(x; \theta_f)$ is extracted, and each label $y_i$ is embedded in a vector $v_i \in \mathbb{R}^{d_l}$ which is jointly learned during training. Context-dependent node vector $h_i = [z;v_i]$ for label $y_i$ is initialized by concatenating the latent input representation $z$ and label vector $v_i$.
A graph feature matrix $H \in \mathbb{R}^{|V| \times (d_z + d_l)}$ is constructed so that the $i$-th row of $H$ is $h_i$, i.e., $H = [h_1, h_2, ..., h_i, ..., h_N]^\top$. 

\paragraph{Tied-Weight Decoder} 
After $L$ iterations of graph convolution, the graph feature $U = H^{(L)}$ is extracted.
The output feature vector $u_i$ of label $y_i$ is decoded by tying output weight between the input latent representation $z$ and node representation $v_i$ \citep{DBLP:journals/corr/InanKS16}. The label score $F_i$ is obtained as $F_i = u_i z + u_i v_i$, where $u_i$ is the $i$-th row of $U$. We constrain $z$ and $v_i$ to have the same dimension.

Unlike the MLP+CRF, the proposed method with neural message passing encodes labels based on vector representations, and therefore potentially encoding richer information. 

\section{Related Work}
\paragraph{Structured Prediction with Label Relations}

Structured prediction has been used for classification with label relations \citep{taskar2004max}, 
\citep{tsochantaridis2005large}, \citep{lampert2011maximum}, \citep{NIPS2012_4520}, \citep{bi2011multi}, \citep{zhang2017hierarchical}.

The goal of our work is clearly distinguished from the aforementioned works. Structured prediction can be viewed as a variant of multi-label classification, it takes input data with multiple assignments during training and jointly predicts a set of class labels for new observations during testing, while in our work the proposed model is trained on single-labeled data.

\paragraph{Classification with Label Relations}

\citet{deng2014large} incorporated WordNet \citep{wordnet} in object recognition and demonstrated that exploiting label relations not only improves multi-class classification accuracy but also multi-label classification performance, by setting hard constraints on the exclusive and inclusive relation between labels. This model was further extended for soft label relations using the Ising model in \citep{DBLP:journals/corr/DingDMN15}. 

There are three major differences between this approach and the proposed approach.
First, we do not impose any constraints on the graph structure other than requiring the availability of pairwise relations among nodes. \citet{deng2014large} on the other hand proposed to used a special kind of representation (the HEX graph) to express and enforce exclusion, inclusion, and overlap relations. 
Second, the proposed model is trained strictly with a single-labeled data, while \citet{deng2014large} and 
\citet{DBLP:journals/corr/DingDMN15} add multiple labels during training by using hard constraints. 
Third, we train and use the entire model as a deterministic network, while \citet{deng2014large} and \citet{DBLP:journals/corr/DingDMN15} require a separate inference procedure to model a conditional probability in the test time, 
leading to mismatch between training and test. 

\paragraph{Graph Convolutional Networks}

Previous works focused on exploiting different GCN structures. For instance, \citet{DBLP:journals/corr/DefferrardBV16} approximated smooth filters in the spectral domain using Chebyshev polynomials with free parameters that are learned in a neural network-like model. \citet{kipf2016semi} 
introduced simplifications that significantly improves both training times and predictive accuracy.

The main difference between the proposed method and the previously mentioned works is in the input data structure. The method proposed applies GCN as a layer in DNN to model data with structured output instead of structured input. In particular, the proposed model projects structured labels into a high-dimensional space and forward label hidden states conditioned on input data to GCN layers for feature extraction {\it and} classification. 

\paragraph{Classification with External Knowledge}

Recent works have begun to investigate new ways to integrate richer knowledge in classification tasks. For example, \citet{grauman2011learning}, \citet{hwang2012semantic} and \citet{deng2014large} took the WordNet category taxonomy to improve image object recognition. \citet{mcauley2012image} and \citet{johnson2015love} used metadata from a social network to improve image classification. \citet{ordonez2013large} leveraged associated image captions to estimate entry-level labels of visual objects.
\citet{Hu_2016_CVPR} used label relation graphs and concept layers for layered predictions.

The proposed method is a novel approach to incorporating external knowledge about label relations and is not task-specific. The label structure can be extracted in an arbitrary way. 

\section{Experiments Settings}

\subsection{Datasets}
The proposed model is assessed in two experiments: a visual object recognition task on a canine image dataset, and a document classification task on an in-house dataset. Dataset statistics are summarized in Table~\ref{table:data}. 

\begin{table}[t]
\centering
\caption{\small Data Statistics}
\vspace{-3mm}
\label{table:data}
\resizebox{\linewidth}{!}{
\begin{tabular}{@{}lrrr@{}}
\toprule
\textbf{Dataset}  & \textbf{Nodes (Labels)} & \textbf{Edges} & \textbf{Data Size} \\ \midrule
Canine Images & 170 & 170 & 23,800 \\ 
In-House Documents & 251 & 15,498 & 28,916 \\\bottomrule
\end{tabular}}
\vspace{-6mm}
\end{table}

\subsubsection{Canine Image Dataset}
The canine image dataset is composed of open-source images with labels on a subgraph of WordNet, which is a hierarchical structure of objects. With this dataset we want to evaluate the performance of the proposed model on a special case of graph structure: a tree structure, where edges are directed and each node has only one edge (apart from the parent node).

\paragraph{Image data collection} 
We collect a new dataset consisting of images using an approach inspired by \citet{evtimova2017emergent}. We crawl the nodes in the subtree of the `canine' synset in WordNet, and query the label of each node in Flickr to retrieve 140 images. Images in each node are partitioned into 100/20/20 images for training/validation/test sets respectively.

\paragraph{Label graph construction} 
The adjacency matrix is extracted from the WordNet canine subgraph. 

\paragraph{Input Representation}
Input features are extracted by ResNet-34 \citep{he2016deep} pre-trained on ImageNet. The 512-dimensional feature vector is extracted after applying average pooling to the 512-channel 8$\times$8 feature maps from the final convolutional layer. We do not fine-tune the network.

\subsubsection{Document Classification Dataset}
We use an in-house dataset composed of various types of web page content for document classification. The underlying output structure is generated using semantic similarity of labels. We use this dataset to evaluate the performance of proposed model on general undirected graph structure.

\paragraph{Data collection and preprocessing} 
In this dataset, each document has one human-annotated label that summarizes the primary information of the content. The labels are topics covering company names, business, finance, accounting, marketing, human resource, technology, lifestyle, and more. The dataset is split by 60\%/20\%/20\% into training/validation/test sets respectively. The documents are lowercased and tokenized.
The vocabulary contains the most frequent 100,000 unigrams and bigrams.

\paragraph{Label graph construction} 
The label graph is built by measuring pairwise label similarities based on the label definitions. 
The label definition is retrieved from Wikipedia.
If the label name does not have an exact match, the top-three topics suggested by Wikipedia are selected, and their Wikipedia definitions are concatenated as an alternative.
The definitions are further tokenized into words and vectorized. We experiment with definition vectors created by TF-IDF weighted average of pre-trained word vectors from \cite{DBLP:journals/corr/JoulinGBM16}.

The adjacency matrix is created by computing the pairwise cosine similarities of definition vectors. The $[i, j]$ entry of an adjacency matrix $A \in \mathbb{R}^{|V|\times|V|}$ is $A_{i,j, i \neq j} = \frac{\mu_i \cdot \mu_j}{\norm{\mu_i}_2 \norm{\mu_j}_2}$,
where $\mu_i$ and $\mu_j$ are the $i$-th and $j$-th label's definition vector respectively.
A discrete adjacency matrix $\bar{A}$ is built by setting a threshold $\epsilon$ on the continuous adjacency matrix.
In the experiments, we set $\epsilon$ as the 75\% percentile of all entries in $A$. The discrete adjacency matrix $\bar{A}$ has on average 29.74 edges per node.

\paragraph{Input Representation}
After preprocessing, the document vectors are embedded in a CBoW manner \citep{mikolov2013efficient}. 
Let $C\in \mathbb{R}^{v \times d_e}$ be a trainable embedding matrix, where $v$ is the vocabulary size and $d_e$ the embedding dimension, and $c_i$ be the of $i$-th row of $C$. The embedding function $E(\cdot)$ for document with tokens indices $D = \{d_1, d_2, ..., d_N\}$ is defined as $E(x) = \frac{1}{N} \sum_{n=1}^N c_{d_n}$.

\subsection{Model and Learning Configuration}

\paragraph{Proposed Model: GCNTD}
We used a model architecture with two layers of GCN propagation and tied-weight decoder. We also experiment with 4 and 8 layers, but did not observe significant difference on the model performance.
\paragraph{Baselines}
We consider the following baselines.
\begin{description}
\itemsep -.3em
  \item[$\bullet$ MLPn]
The baseline model uses $n$ layers of multi-layer perceptron (MLP) instead of the proposed GCN layers to map the contextual hidden state $z$ to the classes. We denote this model as MLPn.
  \item[$\bullet$ MLP-CRF] This baseline is described in section 3.2.
We apply mean-field inference (MF) in our experiment. We fine-tune MLP parameters and $\theta$ on validation set using top-1 accuracy.
\item[$\bullet$ GCNTD-\textit{fc/id}] Another baseline is to set the adjacency matrix in GCNTD as fully-connected (\textit{fc}) or identity matrix (\textit{id}). 
\end{description}

All models are trained by minimizing negative log-likelihood (NLL) with back-propagation using Adam optimizer \citep{kingma2014adam} with an initial learning rate of 0.001. The learning rate is annealed each time the validation error does not improve. Each training is early-stopped based on the top-1 accuracy on the validation set. We random search embedding size $d_e$ and learning rate on validation set \cite{bergstra2012random}. Metrics are reported on a test set using the best model according to validation set. We observed similar training time for all the models.

\subsection{Evaluation}
Apart from the top-1 and top-10 accuracies, we propose several graph-theoretic metrics to understand the performance of the proposed GCNTD model. We refer to `predictions' as the 10 labels that are predicted with the highest scores, if not otherwise specified. The following are our evaluation metrics:
\begin{description}
\itemsep -.3em
  \item[$\bullet$ Top-1/top-10 accuracy] The percentage of test cases when the true label is predicted in top-1/top-10 predictions.
  \item[$\bullet$ One-hop precision@k] The fraction of top-k predictions that overlaps with the true label and its one-hop neighbors. By default $k=10$. 
  \item[$\bullet$ One-hop recall@k] The fraction of the true label and its one-hop neighbors that overlaps with top-k predictions. By default $k=10$. 
  \item[$\bullet$ Top-1/top-10 distance] Distance refers to the shortest path between a certain prediction and the true label on the graph, if they are connected at all. Top-1 distance is the distance between top-1 prediction and true label, and top-10 distance is the average distance between top-10 predictions and true label. 
  \item[$\bullet$ Diameter] The diameter of a graph is the maximum eccentricity of any vertex in the graph. In other words, it is the greatest distance between any pair of vertices. Here the diameter refers to the diameter of the subgraph that the top-k predictions form.
\end{description}

The one-hop precision and recall are similar to those in multi-label classification framework. For simplicity, we refer to them as precision and recall. In these metrics, it is assumed that the true label's one-hop neighbors on the graph are also potential true labels. Let true label and its one-hop neighbors be $T$ and predictions be $P$, the precision and recall are:
$\text{precision} = |T \cap P|/|P|$, $\text{recall} = |T \cap P|/|T|.$
In addition to the ability of model to exactly match the ground truth, these two metrics also measure the ability for the model to find a small cluster around ground truth. Higher values on top-1/top-10 accuracy, precision, and recall indicate stronger predictive power.

Top-1/top-10 distances and diameter, on the other hand, measure the coherence of predictions from a graphical perspective.
Since the graph captures label relations, the labels that are closer to each other on the graph are more related. In the case of the definition-based label graph in document classification task, the graph captures semantic similarities. These metrics hence measure how centralized the predictions are with respect to the true label or themselves from a semantic perspective. Lower values indicate semantically more related labels.

\begin{table}[t]
\centering
\caption{\small Object recognition on canine image dataset with GCNTD and MLP1 (1 layer MLP), evaluated on the testing set with all nodes (\textit{all}) or with only leaf nodes (\textit{leaf}). GCNTD has better performance on graph-theoretic metrics such as top-10 distance and diameter, while MLP achieves higher accuracy. 
}\label{table:cv}
\label{table:cv}

\vspace{-2mm}
\resizebox{\linewidth}{!}{
\begin{tabular}{@{}r|c|lllll@{}}
\toprule
 & \begin{tabular}[c]{@{}c@{}}\textbf{Test}\\ \textbf{data} \end{tabular}&\begin{tabular}[c]{@{}c@{}}\textbf{\small Top-1(10)}\\\textbf{\small accuracy} \end{tabular} & \textbf{\small Precision} & \textbf{\small Recall} & \begin{tabular}[c]{@{}c@{}}\textbf{\small Top-1(10)}\\\textbf{\small distance} \end{tabular} & \textbf{\small Diameter} \\ 
\midrule
GCNTD&\multirow{4}{*}{\begin{tabular}[c]{@{}c@{}}\\all\end{tabular}} & .42 (.74) & .15 & .60 & 2.07  (3.46) & 3.37 \\
MLP1 & &\textbf{.44} (\textbf{.75})& .14 & .57 & 2.02 (3.52) & 3.86 \\
MLP-CRF& &.42 (\textbf{.75})& \textbf{.16} & \textbf{.67} & \textbf{1.97} \textbf{(3.29)} & 4.33 \\
GCNTD-\textit{id}& &.40 (.70) & .13 & .52 &2.17 (3.64)& 5.00 \\
GCNTD-\textit{fc}& &.42 (.74)& .14 & .56 & 2.11 (3.57) & \textbf{2.73} \\
\midrule
GCNTD&\multirow{4}{*}{\begin{tabular}[c]{@{}c@{}}\\leaf\end{tabular}} & .47 (.76)& .13 & .64 & 2.02 (3.57) & 3.26 \\
MLP1 && \textbf{.49} (\textbf{.77}) & .11 & .62 & \textbf{1.95} (3.64) & 3.70 \\
MLP-CRF& &.43 (.74) & \textbf{.14} & \textbf{.70} & 1.95 \textbf{(3.41)} & 4.25 \\
GCNTD-\textit{id}& & .45 (.71) & .11 & .56 & 2.10 (3.77)  & 6.00 \\
GCNTD-\textit{fc} && .48 (.75) & .12 & .60 & 2.02 (3.69) & \textbf{2.42} \\
\bottomrule
\end{tabular}}
\vspace{-5mm}
\end{table}

\begin{figure}[t]
  \begin{minipage}[c]{0.5\linewidth}
    \includegraphics[width=\linewidth]{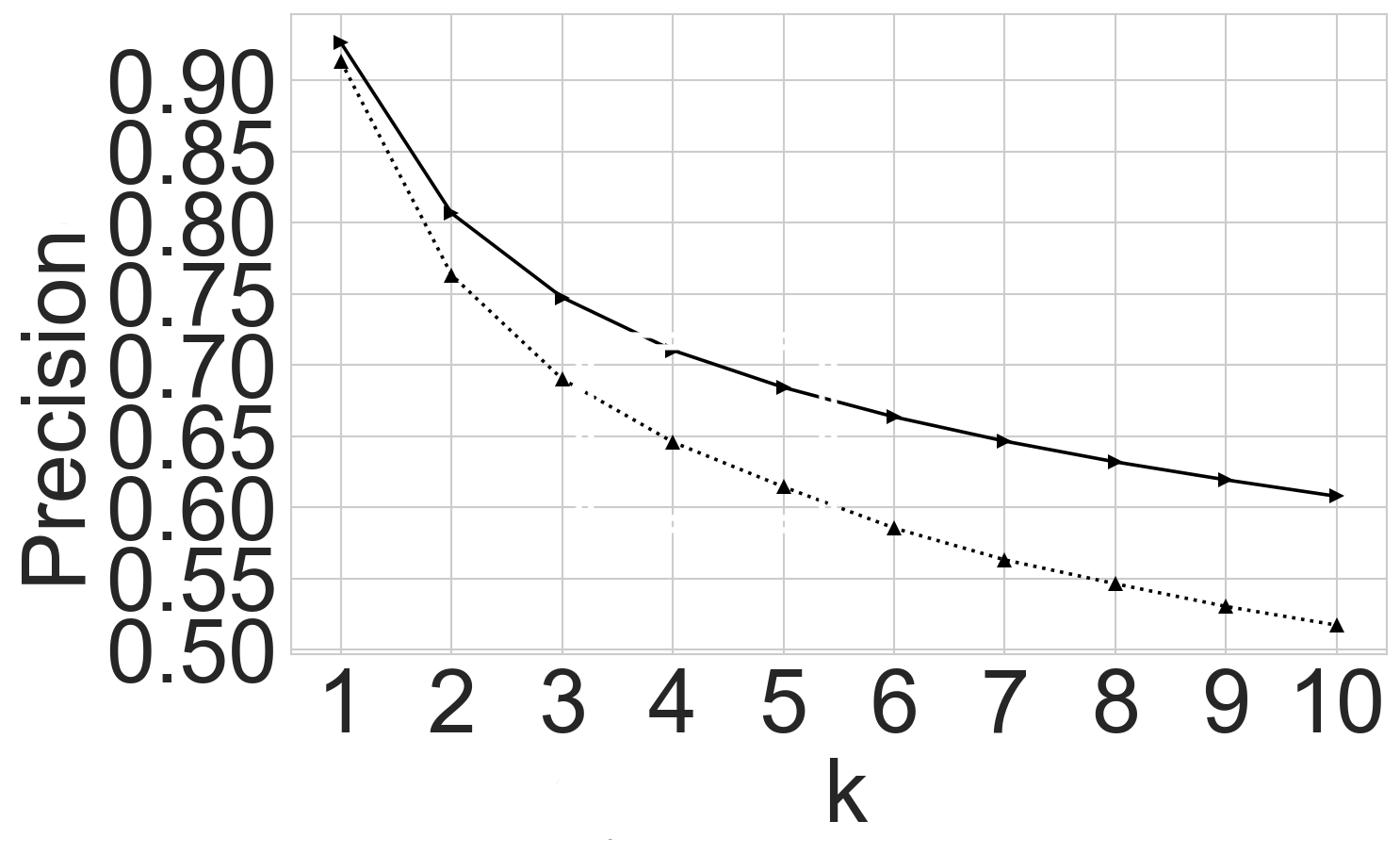}
  \end{minipage}\hfill
  \begin{minipage}[c]{0.5\linewidth}
   \includegraphics[width=\linewidth]{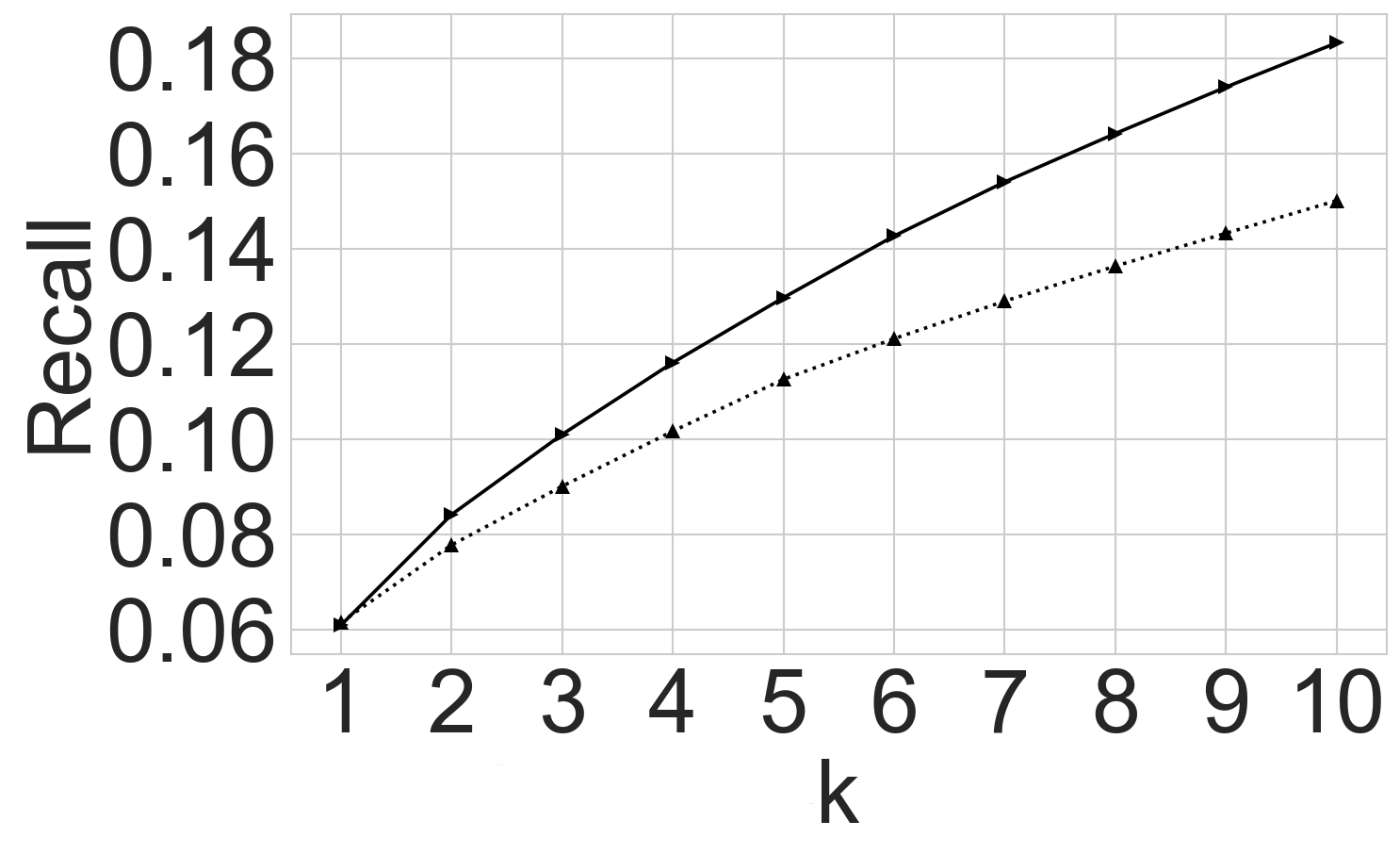}
  \end{minipage}
  \centering
\includegraphics[width=0.5\linewidth]{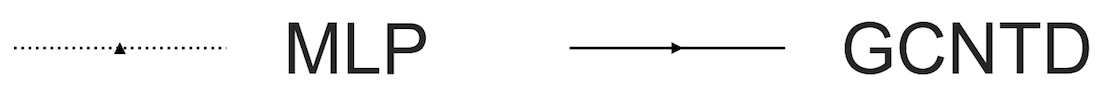}
\vspace{-0.2cm}
\caption{\small Precision and recall@k in document classification. GCNTD refers to average precision/recall of 10 GCNTD in Table \ref{table:cv}, and MLP refers to that of MLP. GCNTD significantly outperforms MLP, and the improvement grows as k increases.}\label{fig:precision}
\vspace{-7mm}
\end{figure}

\section{Results}


\subsection{Object Recognition: Canine Image Dataset}
The results are shown in Table \ref{table:cv}, where the models were trained on all the nodes and evaluated on either all the nodes (\textit{all}) or the leaf nodes only (\textit{leaf}). In both cases, the models that consider the underlying label structure (i.e. GCNTD and MLP-CRF) achieve higher performances on graph-theoretic metrics
while the MLP outperforms the GCNTD and MLP-CRF on accuracy. The MLP-CRF achieves the highest precision and recall on both evaluation scenarios. Such result indicates that the explicitly defined energy functions in a graphical model are often beneficial for predicting labels closer to the ground truth than the MLP and the GCNTD on WordNet hierarchy. 

\begin{table}[t]
\centering
\caption{\small Document classification results of GCNTD and MLP. The GCNTD significantly improves the MLP-CRF and the MLP1, MLP2, MLP4 (1, 2, and 4 layers of MLP) on accuracies, precision, and recall. 
}\label{table:doc}

\vspace{-2mm}
\resizebox{\linewidth}{!}{
\begin{tabular}{@{}r|lllll@{}}
\toprule
 & \begin{tabular}[c]{@{}c@{}}\textbf{\small Top-1(10)}\\ \textbf{\small accuracy} \end{tabular} & \textbf{\small Precision} & \textbf{\small Recall} & \begin{tabular}[c]{@{}c@{}}\textbf{\small Top-1(10)}\\ \textbf{\small distance} \end{tabular} & \textbf{\small Diameter} \\ 
\midrule
GCNTD & .83 (.95) & \textbf{.61} & \textbf{.18} & .24 (1.35) & \textbf{2.40} \\
GCNTD-\textit{id} & .81 (.94)& .50 & .15 & .28 (1.48) & 2.81\\
GCNTD-\textit{fc} & \textbf{.84} (\textbf{.96})& .50& .15 & \textbf{.22} (1.48) & 2.82 \\
MLP-CRF & .81 (.95) & .60& .16 & .25 (\textbf{1.34}) & 2.42 \\
MLP1 & .82 (.95) & .53 & .16 & .26 (1.44) & 2.67 \\
MLP2 & .81 (.94) & .50 & .14 & .27 (1.49) & 2.71 \\
MLP4 & .75 (.91) & .46 & .13 & .37 (1.54) & 2.94 \\
\bottomrule
\end{tabular}
}
\vspace{-5mm}
\end{table}

\begin{figure}[t]
  \begin{minipage}[c]{0.5\linewidth}
    \includegraphics[width=\linewidth]{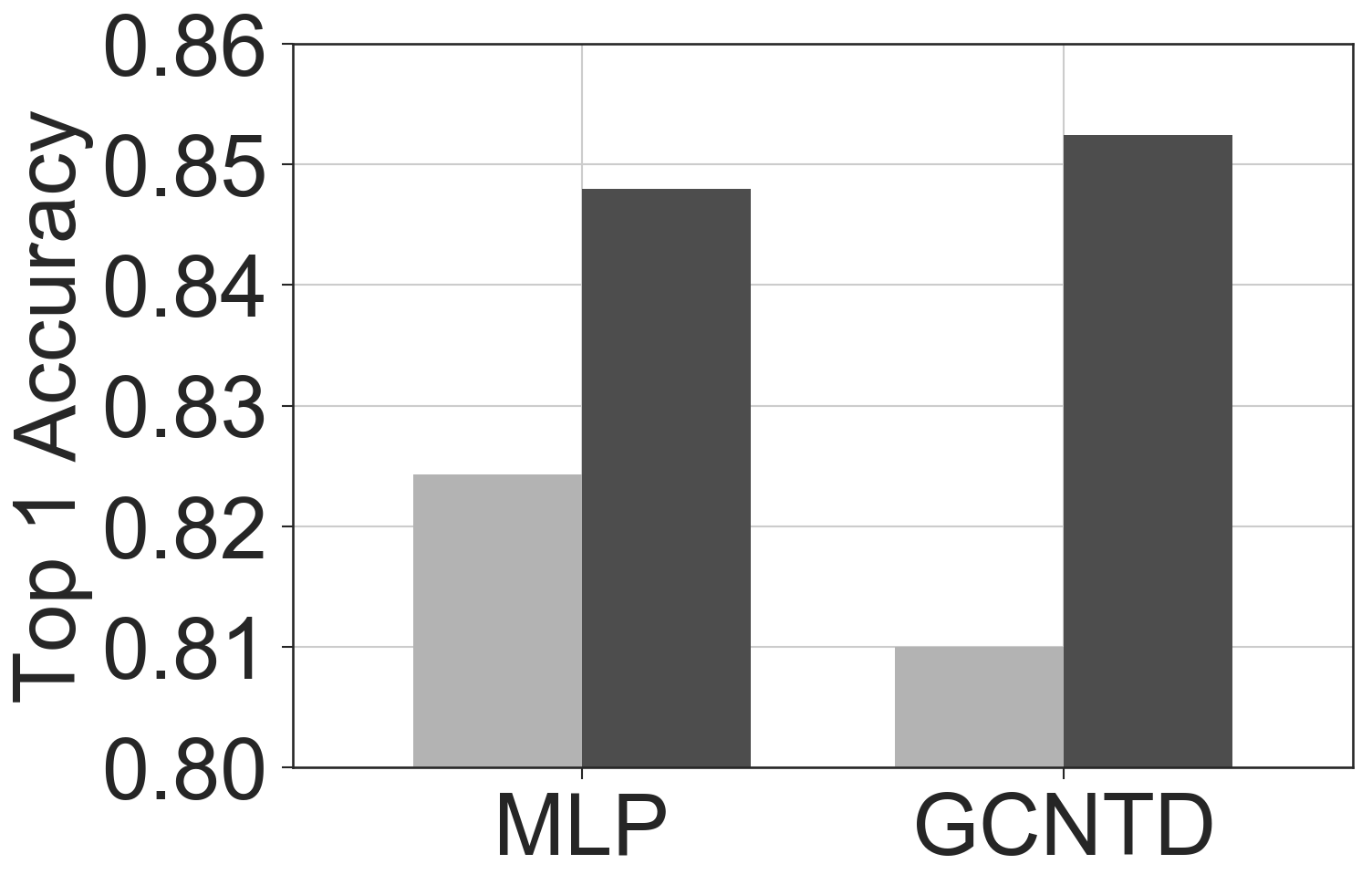}
  \end{minipage}\hfill
  \begin{minipage}[c]{0.5\linewidth}
   \includegraphics[width=\linewidth]{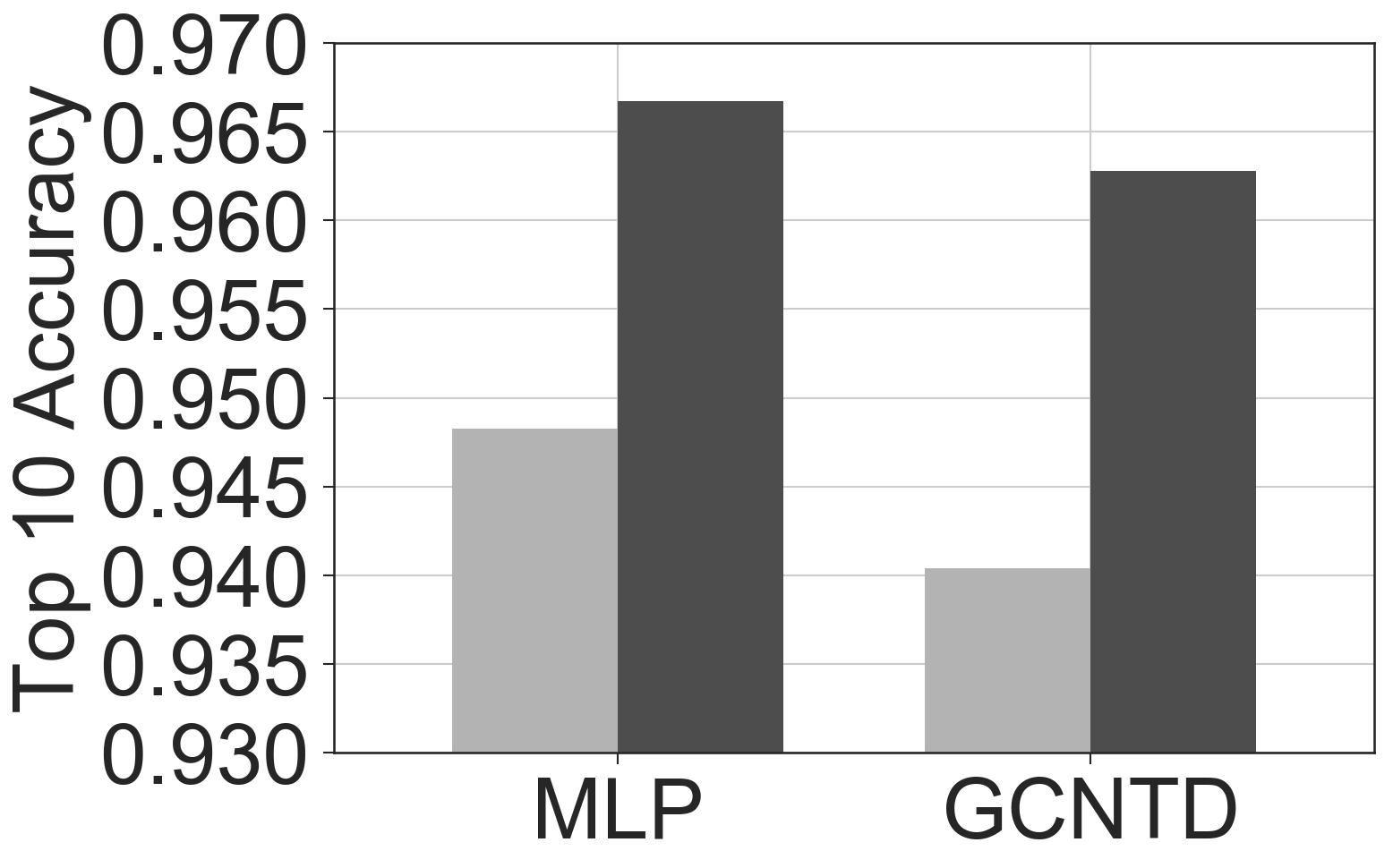}
  \end{minipage}
  \centering
\includegraphics[width=0.6\linewidth]{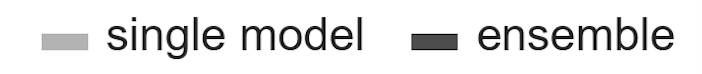}
\vspace{-0.2cm}
\caption{\small The impact of ensemble method on top-1/top-10 accuracy in document classification. The single model result refer to the average accuracy of ten single models with identical hyperparameter settings. Ensemble refers to the accuracy of an ensemble of such 10 models. Ensemble has a stronger boosting effect on GCNTDs than MLPs on both top-1 and top-10 accuracies.}\label{fig:ensemble}
\vspace{-7mm}
\end{figure}

\subsection{Document Classification}
Results are shown in Table \ref{table:doc}. In general, the GCNTD outperforms the MLP on all the metrics, and outperforms MLP-CRF on accuracies, precision, and recall. GCNTD-\textit{fc} achieves the highest accuracies, indicating the benefit of message passing under the extreme case where labels are fully-connected.

Figure \ref{fig:precision} shows the comparison between GCNTD and MLP in top-k precision and recall .
This demonstrates that the GCNTD tends to find a smaller cluster of predictions that are closer to ground truth. Since the label graph, in this case, is constructed by measuring definition similarities, the GCNTD can be thought of as making predictions that are semantically closer and more related to the ground truth: a smaller diameter indicates semantically closer predictions, and a smaller distance indicates that the predictions are semantically closer to the ground truth.

\section{Discussion and Conclusion}
We have proposed a graph convolutional network (GCN) augmented neural network classifier to exploit an underlying graph structure of labels. The proposed approach resembles an approximate inference procedure in probabilistic graphical models, but replaces iterative inference with graph convolution layers. 
In the experiments on object recognition and document classification, the proposed model achieved better performance on graph-theoretic metrics than a baseline model that ignores label structures. 
The proposed approach can be applied to any classification task with a label graph to improve accuracy and top predictions relevancy. It can be also used to incorporate external knowledge about labels by encoding such knowledge in label graph.

\section*{Acknowledgments}
Meihao Chen and Zhuoru Lin thank Nicholaus Halecky, Jeffery Payne, Oleg Khavronin, Patrick Kelley, and Lindsay Reynolds for discussion and support on this work. Kyunghyun Cho thanks support by Tencent, eBay, Facebook, Google and NVIDIA, and was partly supported by Samsung Advanced Institute of Technology (Next Generation Deep Learning: from pattern recognition to AI).
This work is completed during Zhuoru Lin's internship with Bombora Inc.

\bibliography{naaclhlt2018}
\bibliographystyle{acl_natbib}

\end{document}